# A Regularization Method to Improve Adversarial Robustness of Neural Networks for ECG Signal Classification

Linhai Ma, Liang Liang

*Abstract*— Electrocardiogram (ECG) is the most widely used diagnostic tool to monitor the condition of the human heart. By using deep neural networks (DNNs), interpretation of ECG signals can be fully automated for the identification of potential abnormalities in a patient's heart in a fraction of a second. Studies have shown that given a sufficiently large amount of training data, DNN accuracy for ECG classification could reach human-expert cardiologist level. However, despite of the excellent performance in classification accuracy, DNNs are highly vulnerable to adversarial noises that are subtle changes in the input of a DNN and may lead to a wrong class-label prediction. It is challenging and essential to improve robustness of DNNs against adversarial noises, which are a threat to life-critical applications. In this work, we proposed a regularization method to improve DNN robustness from the perspective of noise-to-signal ratio (NSR) for the application of ECG signal classification. We evaluated our method on PhysioNet MIT-BIH dataset and CPSC2018 ECG dataset, and the results show that our method can substantially enhance DNN robustness against adversarial noises generated from adversarial attacks, with a minimal change in accuracy on clean data.

Linhai Ma is now with the Department of Computer Science, University of Miami, Coral Gables, FL 33146 USA (e-mail: l.ma@miami.edu).
Liang Liang is now with the Department of Computer Science, University of Miami, Coral Gables, FL 33146 USA (e-mail: liang@cs.miami.edu).

## I. INTRODUCTION

Electrocardiogram (ECG) is often the first step for the diagnosis of heart conditions before using expensive cardiac medical imaging, and it is widely used for long-term monitoring of a patient with chronic heart disease. To become an expert for the interpretation of ECG recordings, in-depth knowledge and medical training on human physiology and cardiology are needed. For patients in the developing countries where human-expert cardiologists are lacking, computational methods, such as machine learning methods, may provide an accessible and affordable solution for automated ECG signal analysis, e.g., classifying ECG signals into different disease categories. Among the machine learning techniques, deep neural networks (DNNs) including convolutional neural networks (CNNs), which have demonstrated superior performance in a wide range of applications [1], were successfully applied for ECG signal classification [2-4]. For example, an adaptive one-dimensional (1D) CNN for ECG classification on the MIT-BIH dataset was designed by S. Kiranyaz et al. in 2015 [5], which could be the first study of using CNN for ECG signal classification. U. Rajendra et al. [6] developed a deep CNN to classify the ECG signals, and their model achieved an accuracy of 94.03% on the MIT-BIH dataset. M. Kachuee et al. [7] designed a CNN with residual connections, which achieved an accuracy of 93.4%. on the MIT-BIH datasets. Awni Y. Hannun, et al. [8] developed a CNN to classify 12 rhythm classes using a dataset of 91,232 single-lead ECGs from 53,877 patients, which is the largest, yet private dataset compared to other public datasets, and the classification accuracy was similar to that of human-expert cardiologists. Thus, with a sufficiently large amount of data and a carefully designed neural network structure, a well-trained DNN model could reach the human-expert level for ECG signal classification. For a complete review of DNNs for ECG signal analysis, we refer the reader to the survey [4], which summarized nearly 200 papers from 2015.

It seems that artificial intelligence (AI) doctors are technically ready to serve the patients for ECG diagnosis. However, despite the high classification accuracy, DNNs are known to be vulnerable to adversarial noises in the field of computer vision [9], which are small perturbations to the input of a DNN model, even imperceptible to human eyes, and the adversarial noises can change the prediction output of the DNN model. Adversarial noises are often generated by adversarial attack algorithms in two categories: white-box or black-box attacks. It is a white-box adversarial attack if the attacker knows the inner structure (including weights) of the DNN. White-box adversarial attacks often use the gradient information (e.g., the gradient of the output with respect to the input) of the target DNN to construct adversarial noises and add the noises to the input of the DNN. Two well-known white-box attacks are Fast Gradient Signed Method (FGSM) [10] and Projected Gradient Descent (PGD) [11]. It is a black-box attack if the attacker has no knowledge of DNN inner structure and can only query the DNN to get the outputs. The typical approach for black-box attacks is transfer-based, in which a substitute DNN model is trained on the input-output pairs of the target DNN model (called oracle) to be attacked, so that the substitute model can learn the decision boundary of the oracle; after training, the substitute is used to construct adversarial noises to attack the oracle [12]. The weakness of transfer-based attacks is studied by Jonathan Uesato et al. [13], which shows that the success of the attack is highly dependent on the similarity between the substitute and the oracle. Another approach for black-box attacks is using stochastic sampling methods to estimate the gradient of the target network, and then the estimated gradient can be used for attacks [13], but this approach has high computation cost and therefore is very time-consuming. From the perspective of the attacker, black-box attacks are more difficult to implement and less likely to succeed, compared to white-box attacks. From the perspective of the defender, in this study, we consider the worst-case scenario that the attacker knows the DNN structure (i.e., white-box attacks). Adversarial attacks pose significant threats [9] to the DNN-based systems in sensitive and life-critical application fields, e.g., vision-based self-driving. Recently, a special white-box attack, named smooth adversarial perturbations (SAP) is designed to fool ECG signal classifiers [14].

To improve DNN robustness against adversarial noises for image classification applications, lots of effort has been made to develop defense methods in the field of computer vision [13, 15, 16]. Currently, the most popular and effective defense strategy is adversarial training, and the basic idea is to generate adversarial noises through an adversarial attack, add the noises to the images, and then use both the clean images (without adversarial noises) and the noisy images (added with adversarial noises) for training a DNN model. A noisy image is called an adversarial sample. Through adversarial training, a DNN model may learn features of adversarial samples, and then its decision boundary for classification is updated so that it will become difficult to "push" the input samples/images across the decision boundary by adding a small amount of adversarial noise. Different adversarial training-based defense methods have been proposed [10, 11, 17-19], which share the same basic idea and vary in how the adversarial samples are generated. Adversarial training is straightforward but not perfect. For example, it is very computationally expensive and time-consuming to generate adversarial samples, and very-noisy adversarial samples can be misleading and reduce the classification accuracy of the DNNs.

Parallel to adversarial training, regularization terms can be added to the loss function during model training to reduce the sensitivity of the model output with respect to the input.

Regularization terms could be the gradient magnitude of the loss [20] or Jacobian regularization [21] which has good performance for image classification applications. However, little effort has been devoted to solving the robustness issue in ECG applications. Also, the robustness issue is unfortunately neglected by the US Food & Drugs Administration (FDA): some AI/DNN algorithms for ECG analysis [22] have already been FDA-cleared/approved, making impacts on human lives. For the best interest of the consumers/patients, it is urgently needed to develop methods to improve DNN robustness against adversarial noises.

In this paper, we present a novel method to improve DNN robustness against adversarial noises by minimizing the noise-to-signal ratio (NSR) during the training process. We used the PGD algorithm [11] to implement the white-box adversarial attacks, which is widely used for robustness evaluation [15]. In addition, we tested the recent SAP adversarial attack [14] specifically designed for ECG signals. We evaluated the proposed method on two public datasets: PhysioNet MIT-BIH Arrhythmia ECG (MIT-BIH) dataset [23] and China Physiological Signal Challenge 2018 (CPSC2018) [24, 25] dataset, and the results show that our proposed method leads to an enhancement in robustness against the adversarial attacks, with a minimal reduction in accuracy on clean data. The code of our method is publicly available.

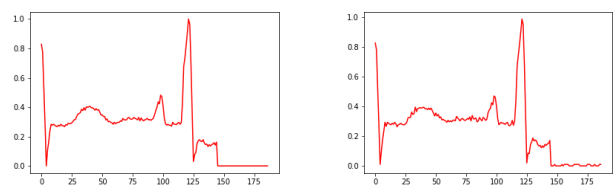

(a) classify clean ECG as "N"    (b) classify "noisy" ECG as "V"
Fig. 1. a ECG single (a) was added with adversarial noise (b)

Before diving into the details of the method, we would like to show the severity of the issue by an example in Fig. 1. Given an ECG signal $x$ (we call it clean signal although it may have some noises during the ECG recording process) as input, a DNN model will predict the class label of the signal. The true class label is "N", and the prediction from the model is correct for the input $x$ (Fig.1(a)). A noisy signal $x_\epsilon = x + \delta$ is obtained (Fig.1(b)) by adding a small amount of adversarial noise $\delta$ to $x$ through an adversarial attack. Given $x_\epsilon$ as input, the DNN model outputs the class label "V". The amplitude $\epsilon$ of the noise $\delta$ is so small that it is hardly detectable by the human eyes, but the DNN model thinks that $x$ and $x_\epsilon$ are completely different signals and therefore assigns different class labels. Would you trust an AI-based ECG-analysis system if it may make such "unbelievable" mistakes?

## II. METHODOLOGY

### A. The Proposed Method using NSR Regularization

In this section, we will introduce a new loss function to improve DNN robustness by reducing noise-to-signal ratio (NSR). The noise refers to the adversarial noise generated by an adversarial attack. The signal refers to the output of the DNN classifier (i.e., the logits before the softmax layer). Our method assumes that the nonlinear activation function is ReLU or its variants (e.g., LeakyReLU), and the DNN may have the standard layers, including convolution layers, fully connected layers, pooling layers, batch-normalization layers, dropout layers, and skip-connections. Since a convolution layer is equivalent to a fully connected layer with shared weights, a CNN can be converted into a multiple layer perceptron (MLP) in theory. Therefore, we will introduce our method from the perspective of MLP. The idea is to apply regularization with the goal of minimizing NSR. Given an input sample/signal $x$ (reshaped to a vector), the output of a neural network can be exactly expressed by a linear equation [18]:

$$z = W^T x + b \tag{1}$$

where the weight matrix $W$ is an implicit function of $x$, and the bias vector $b$ is also an implicit function of $x$. Thus, the network is a piecewise linear function on the input space. The output of the network is a vector $z = [z_1, \dots, z_K]^T$, where $z_i$ is the output logit of class-$i$, and the total number of classes is $K$. Let $w_i$ be the $i$-th column of $W$ and $b_i$ be the $i$-th element of $b$. Let $y$ be the true label of $x$. Then we have:

$$z_y = w_y^T x + b_y \tag{2}$$

During an adversarial attack, a noisy $\delta$ (a vector) is generated and added to the input $x$, and then the new output is:

$$z_{y,\delta} = w_{y,\delta}^T (x + \delta) + b_{y,\delta} \tag{3}$$

By definition [26], adversarial noise is small but may lead to the change of the output. When the amplitude of the noise $\delta$ is small enough, then $w_y \approx w_{y,\delta}$ and $b_y \approx b_{y,\delta}$. Therefore, we obtain:

$$z_{y,\delta} \approx w_y^T x + b_y + w_y^T \delta = z_y + w_y^T \delta \tag{4}$$

We define NSR and apply Hölder's inequality:

$$NSR_y = \frac{|w_y^T \delta|}{|z_y|} \leq \frac{\|w_y\|_q \cdot \|\delta\|_p}{|z_y|} \tag{5}$$

where $\frac{1}{p} + \frac{1}{q} = 1$. In this work, we focus on L-infinity norm $\|\delta\|_\infty = \epsilon_\delta$, and therefore:

$$NSR_y \leq \frac{\|w_y\|_1 \cdot \epsilon_\delta}{|z_y|} = R \tag{6}$$

By combining the regularization term $R$ with mean square error (MSE) loss and margin loss for classification, the complete loss function for training the model is obtained:

$$L_{NSR} = (z_y - 1)^2 + \sum_{i \neq y}(z_i - 0)^2 + \sum_{i \neq y} max(0, 1 - z_y + z_i) + \beta \log(R + 1) \tag{7}$$

In the experiment, $\epsilon_\delta$ is set to 1 because the maximum signal amplitude is 1, and β is determined on the validation set. The margin loss $\sum_{i \neq y} max(0, 1 - z_y + z_i)$ and the regularization term $\log(R + 1)$ are only used for correctly classified samples; and for wrongly classified samples, the $L_{NSR}$ only contains the MSE loss $(z_y - 1)^2$. For the experiment on the CPSC2018 dataset, to facilitate the convergence during the training process, the regularization term and the margin loss will not be added to the loss function until the 11-th epoch.

To use the NSR Regularization-based new loss function defined by (7), a proper β is needed to achieve a good balance between robustness and accuracy. For this purpose, we define a metric to measure the overall performance of a classifier, which is given by:

$$ACC_{robust} = \sqrt{ACC_{clean} \times AUC_{\|\delta\|_\infty \leq \epsilon_{max}}} \tag{8}$$

Here, $ACC_{clean}$ is the classification accuracy on clean data (without adversarial noises). $AUC_{\|\delta\|_\infty \leq \epsilon_{max}}$ is the normalized area under the curve of classification accuracy vs noise level, which is the average classification accuracy computed across a range of noise levels, i.e., $\|\delta\|_\infty \leq \epsilon_{max}$. Therefore, $ACC_{robust}$ is the overall performance measure that takes both clean accuracy (i.e., accuracy on clean data) and robustness (i.e., accuracy on noisy data) into consideration, and a better β should lead to a higher $ACC_{robust}$. By definition [26], adversarial noises are small changes in the input that may result in a large change in the output of a model. Thus, we only need to consider noises within a certain range, i.e., setting a maximum noise level $\epsilon_{max}$ for method evaluation. When the noise amplitude is large enough, the content/meaning of the input signal $x + \delta$ is significantly altered, and the output of the model is expected to change. In the experiments, the $\epsilon_{max}$ of $ACC_{robust}$ for MIT-BIT is 0.1 and the $\epsilon_{max}$ of $ACC_{robust}$ for CPSC is 0.01, and the explanation is in Section III-B.

### B. PGD-based Adversarial Attack and Adversarial Training

For method evaluation, we used the algorithm, named Projected Gradient Descent (PGD), to implement the adversarial attack [11], which is regarded as the strongest first-order white box adversarial attack and is widely used for robustness evaluation [13, 15]. For the convenience of the reader, we provide a brief summary of the PGD-based adversarial attack.

The key equation of the PGD attack is given by:

$$x^k = \prod \left( x^{k-1} + \alpha \cdot sign(\nabla_x J(x^{k-1})) \right) \tag{9}$$

where α is the step size and $x^k$ is the adversarial example from the $k$-th iteration. The adversarial noise after the $k$-th iteration is $\delta = x^k - x$. If the noise $\delta$ is larger than the noise level ϵ of the attack, the projection operation $\prod(\ )$ will project it back onto a ϵ-ball, i.e, $\|x^k - x\|_\infty = \|\delta\|_\infty \leq \epsilon$. Assuming $x$ is correctly classified by a DNN classifier, then, after a total number of $K$ iterations, the objective $J$ of the attack on the noisy sample $x^K$ becomes very large such that $x^K$ is wrongly classified. Usually, the objective $J$ is the cross-entropy function that is high when classification accuracy is low.

Therefore, the PGD attack can be summarized by a single function: $x_\epsilon = PGD(x, y, \epsilon, \alpha, K)$, where $x_\epsilon = x^K$ is the generated adversarial sample on the noise level ϵ. The default value of the step size α is 0.01 [11]. The attack strength is determined by $K$, the number of iterations. A PGD attack with $K$ iterations is known as K-PGD, and the 100-PGD is usually used for robustness evaluation [13, 15]. By varying the noise level ϵ, the robustness of a model can be evaluated by measuring its classification accuracy on the adversarial samples in a large range of noise levels.

The basic idea of adversarial training [10] is to generate adversarial samples by an algorithm (e.g., PGD) and use the adversarial samples as part of the training data. In this way, the trained model may become robust against adversarial noises. The loss function of the standard adversarial training [10] is given by:

$$L_{adv} = 0.5 L_{CE}(x,y) + 0.5 L_{CE}(x_\epsilon, y) \quad (10)$$

where $L_{CE}$ is the cross-entropy loss and $x_\epsilon$ is an adversarial sample (e.g., generated by PGD) on the noise level ϵ. As shown in the Section III, the performance of PGD-based adversarial training is determined by the noise level ϵ. To choose a proper value of ϵ, we tried a series of experiments with different values of ϵ, and chose the best one associated with the highest $ACC_{robust}$ ((8)) on validation set.

The standard adversarial training encountered a convergence problem on the CPSC2018 dataset. To help convergence, the adversarial loss $L_{CE}(x_\epsilon, y)$ was not used until the 11-th epoch, and the level of noise is scaling up linearly as the number of epochs increases during training [17]:

$$\epsilon_t = \epsilon(t - 10)/(t_{max} - 10) \quad (11)$$

where $t$ is the index of a training epoch, $t_{max}$ is the total number of training epochs, and $\epsilon_t$ is the level of noise added to the clean samples during the current epoch $t$.

### C. The Jacobian Regularization based Method

For method comparison, we implemented the Jacobian Regularization based method [21] which was designed for image classification applications. The basic idea of the method is to penalize large gradient of the output with respect to the input, which makes the DNN model less sensitive to the change in the input. The sensitivity to input change is measured by the Frobenius norm of the DNN's Jacobian matrix, which can be added to the loss function as a regularization term. The loss function of the method is given by:

$$L_{jacob} = \sum_{n=1}^{N} L_{CE}(x_n, y_n) + \lambda \sqrt{\sum_{d=1}^{D} \sum_{k=1}^{K} \sum_{n=1}^{N} \left(\frac{\partial z_k(x_n)}{\partial x_d}\right)^2} \quad (12)$$

Here, $L_{CE}$ is the cross-entropy loss. $D$ is the input dimension. $K$ is the number of classes. $N$ is the batch size during training. $\{x_1, x_2, \ldots, x_n, \ldots, x_N\}$ are the samples/signals in a mini-batch, and $n$ is sample index in the mini-batch. $y_n$ is the ground truth label of $x_n$. $z = [z_1, \ldots, z_K]^T$ is the output logits of the DNN, and $z_k(x_n)$ is a logit of the input sample $x_n$. $x_d$ refers to the d-th dimension of a sample $x_n$.

The original loss function in (11) caused a numerical problem (inf, nan) for the high dimension data in the CPSC2018 dataset. Thus, we slightly modified the loss function used in our experiment on the CPSC2018 dataset, which is given by:

$$L_{jacob} = \frac{1}{N} \sum_{n=1}^{N} L_{CE}(x_n, y_n) + \frac{\lambda}{NK} \sqrt{\sum_{d=1}^{D} \sum_{k=1}^{K} \sum_{n=1}^{N} \left(\frac{\partial z_k(x_n)}{\partial x_d}\right)^2} \quad (13)$$

To further improve convergence, the regularization term will not be added to the loss function until the 11-th epoch during training on the CPSC2018 dataset.

### D. The Customized DNNs

We tested two DNNs on the MIT-BIH dataset: a CNN and an MLP. The CNN is from a previous work by Kachuee et al. in [7], which has 30 layers in total and achieved excellent classification performance. The MLP is designed by us as a baseline model, and it has 8 layers. The structure of this MLP is (187-128)-RELU-(128-128)-RELU-(128-128)-RELU-(128-32)-(32-5). We refer the reader to our source code for details.

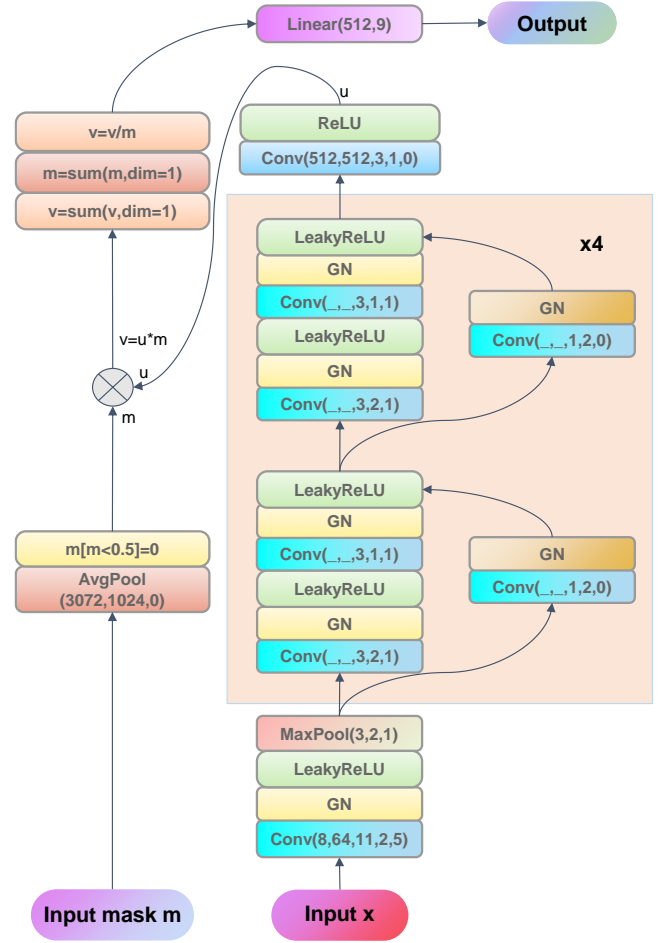

Fig. 2. The high-level architecture of the CNN. The size of the input mask is the same as the size of the input x. The mask will help the network ignore the padded zero elements in x, which has a variable length. The network outputs the classification scores (a.k.a. logits).

For the CPSC2018 dataset, we designed a CNN to classify the ECG signals and handle the challenge of variable input signal length. The high-level architecture of this neural network is shown in Fig. 2. To describe the CNN, we use "Cov (a, b, c, d, e)" to represent a convolution layer with "a" input channels, "b" output channels, kernel size of "c", stride of "d" and padding of "e"; "Linear (f, g)" to denote a linear layer with input size of "f" and output size of "g"; "MaxPool (h, i, j)" to denote max-pooling with kernel size of h, stride of i and padding size of j; "AvgPool (k, l, m)" to denote average-pooling with kernel size of k, stride of l and padding size of m; "GN" to denote group normalization [27]. The network has 4 convolution blocks of the same structure, which will shrink the size of the input tensor and double the number of channels. To handle a time sequence with variable length, RNN is typically used. However, it is known that RNN runs very slow due to the difficulty of parallel computing. To enable the CNN to handle variable signal length, we used zero-padding and input/output-masking. Each signal (i.e., the ECG recording from a patient) is padded with zeros to a fixed length of 33792, so that the padded signals can be assembled into mini-batches for parallel computing. The CNN should ignore the zeros in signals, and this is realized by using an input mask with a fixed length of 33792. For valid elements of the padded signal, the corresponding elements of the mask are ones. For zero elements of the padded signal, the corresponding elements of the mask are zeros. To obtain the output mask, an average-pooling is applied to the input mask by estimating what the mask size should be after a series of convolutional operations on the mask. Then, a valid output feature is obtained by multiplying the output mask and the feature output from the last convolution block. After this output-masking operation, a channel-wise averaging weighted by the mask is performed to reduce the dimension of the feature from a variable length to a fixed length of 512. After the final linear layer, the classification scores (a.k.a. logits) are obtained. We refer the reader to the source code for more details of the CNN model.

## III. Experimental Evaluation of the Methods

### A. Datasets

We used the processed MIT-BIH dataset from a previous study [7, 28], which used annotations in the original MIT-BIH dataset to create five different beat categories in accordance with Association for the Advancement of Medical Instrumentation (AAMI) EC57 standard [29]. The original MIT-BIH dataset consists of ECG recordings at the sampling rate of 360Hz, which involves 47 subjects: 25 men aged between 32 and 89, and 22 women aged between 23 and 89. The recordings were independently labelled beat-by-beat by two or more cardiologists. The processed dataset contains individual heartbeats that have been segmented from the recordings, and subsequently resampled and zero-padded. The ECG signal of one heartbeat is a 1D array of 187 numbers, which is called a sample in this study. The task is to predict the class label of a sample from the 5 categories, which are N (Normal, Left/Right bundle branch block, Atrial escape, and Nodal escape), S (Atrial premature, Aberrant atrial premature, Nodal premature and Supra-ventricular premature), V (Premature ventricular contraction and Ventricular escape), F (Fusion of ventricular and normal) and Q (Paced, Fusion of paced and normal and Unclassifiable). The processed dataset has been divided into a training set (87554 samples) and a test set (21892 samples) in the previous study [7], which is publicly available at [28]. We note that a different train-test split may lead to a different classification accuracy, which makes it difficult to directly compare the ECG classification methods in the literature [4]. By using a publicly available dataset with a given train-test split, the results in this study will become available for direct comparison with new methods that may be developed in the future to improve DNN robustness. To handle class imbalance, we performed up-sampling to ensure that the classes have equal number of samples in the training set and testing set, assuming that the five classes are equally important. We further divided the training set into a "pure" training set (70043 samples, 80%) and a validation set (17511 samples, 20%) that was used to find the best parameters of the methods.

The other dataset, CPSC2018 is described in detail by [25]. There are 6877 ECG 12-lead recordings in this dataset. The ECG signals were sampled by a frequency of 500 Hz, and the duration is in the range of 6 seconds to 60 seconds (note: a few signals have 144 seconds). Namely, each ECG sample (referring to the whole digit ECG signal recording from a patient) in the dataset has 12 leads/channels, and each sample has a variable length (i.e., timepoints) from 3000 to 72000. This dataset has 9 categories, which are "Normal" (918 samples), "Atrial fibrillation (AF)" (1098 samples), "First-degree atrioventricular block (I-AVB)" (704 samples), "Left bundle branch block (LBBB)" (207 samples), "Right bundle branch block (RBBB)" (1695 samples), "Premature atrial contraction (PAC)" (556 samples), "Premature ventricular contraction (PVC)" (672 samples), "ST-segment depression (STD)" (825 samples) and "ST-segment elevated (STE)" (202 samples). There are only class labels on the recording level, and beat-by-beat labels are not available, not like the MIT-BIH dataset that has been labeled beat-by-beat. And therefore, the task is to predict the class label of a recording from a patient. We preprocessed the data according to the examples provided by CPSC2018, and we split the dataset into a training set of 5905 samples, a validation set of 45 samples, and a test set of 450 samples. For each sample, we scaled each lead by its maximum absolute value so that each lead is within the range of -1 to 1 [31], which is the standard way of feature normalization. The details of data preparation (preprocessing, splitting, and augmentation) are in the Appendix.

### B. Model Training and Parameter Tuning

The metric in equation (8) is used for parameter selection. In this study, we set $\epsilon_{max} = 0.1$ for the MIT-BIH dataset and $\epsilon_{max} = 0.01$ for the CPSC2018 dataset. On the MIT-BIH dataset, when the noise level of the PGD-based adversarial attack is higher than 0.1, the signals become very noisy and hardly recognizable by the human eye. On the CPSC2018 dataset, when the noise level is higher than 0.01, the noises in the signals become noticeable by the human eye, and the performance of the methods starts to drop quickly.

Given a dataset and a DNN structure, tuning the parameter of a method is to train DNN models with different parameters on the training set, and measure the performance of each trained model on the validation set. The parameter associated with the best model is the best parameter. For clarity, we give different names to the DNN models trained by the three different methods. A model (CNN or MLP) trained by our new loss $L_{NSR}$ is named "βNSR" where β is the parameter. A model trained by the loss $L_{jacob}$ is named "λJacob" where λ is the parameter. The model trained by the loss $L_{adv}$ is named "adv ϵ" where ϵ is the noise level for adversarial training. In addition, we trained models with the standard cross entropy loss, and the name of each trained model is "CE".

For the experiments on the MIT-BIH dataset, the models were trained for 50 epochs with batch size of 128. The Adamax optimizer was used with the default learning rate of 0.001. For the experiments on the CPSC2018 dataset, the models were trained for 70 epochs with batch size of 64. The Adam optimizer was used with the default learning rate of 0.001.

For our method, we varied the parameter β in the range of 0.1 to 1.0 for MIT-BIT dataset and 0.4 to 1.2 for CPSC2018 dataset, which resulted in a large variation of model performance. The results are reported in Table 1 in Appendix. The best β is 0.4 for the MLP on MIT-BIH validation set with $ACC_{robust} = 0.854$. The best β is 0.3 for the CNN on MIT-BIH validation set with $ACC_{robust} = 0.894$. The best β is 1.0 for the CNN on CPSC2018 validation set with $ACC_{robust} = 0.889$.

For the Jacobian Regularization based method, we varied the parameter λ in a large range, which resulted in a large variation of model performance. The results from the experiments are reported in Table 2 in Appendix. The best λ is 0.9 for the MLP on MIT-BIH validation set with $ACC_{robust} = 0.770$. The best λ is 0.7 for the CNN on MIT-BIH validation set with $ACC_{robust} = 0.806$. The best λ is 24.0 for the CNN on CPSC2018 validation set with $ACC_{robust} = 0.836$.

For the adversarial training method, we varied the parameter ϵ in the range of 0.05 to 0.3 for MIT-BIT dataset and 0.005 to 0.1 for CPSC2018 dataset, which resulted in a large variation of model performance. The results are reported in Table 3 in Appendix. The best ϵ is 0.1 for the MLP on MIT-BIH validation set with $ACC_{robust} = 0.799$. The best ϵ is 0.1 for the CNN on MIT-BIH validation set with $ACC_{robust} = 0.863$. The best ϵ is 0.01 for the CNN on CPSC2018 validation set with $ACC_{robust} = 0.796$.

We used 10-PGD for adversarial training on the MIT-BIH dataset, and 20-PGD for adversarial training on the CPSC2018 dataset. 10-PGD and 20-PGD are weaker than 100-PGD used for method evaluation. We did not use 100-PGD for adversarial training because: (1) it would be cheating to use the same settings for adversarial training and evaluation; and (2) it will take 60 hours to complete the training of a model using 100-PGD with a given parameter, and therefore it will take 660 hours (27.5 days, non-stop) to complete the experiments for finding the best parameter on the CPSC2018 dataset using a Nvidia V100 GPU, and as a result, the total computing cost will exceed our computing budget. Thus, we chose 10-PGD and 20-PGD to keep a feasible computing cost for adversarial training, which is roughly the same as the cost of our method and the cost of the Jacobian Regularization based method.

### C. Evaluation of the Methods using the 100-PGD Attack

For each model trained by a method with the optimal parameter on a dataset (MIT-BIH or CPSC2018), the performance of the model on the test set was evaluated by using the 100-PGD attack. To evaluate a trained model, for each clean sample in a test set, the corresponding adversarial samples on different noise levels were generated by the 100-PGD attack, and the model was applied to classify these adversarial samples to obtain the classification accuracy on different noise levels.

Therefore, for a trained model, a curve of classification accuracy (Acc) vs. noise level is obtained. Besides classification accuracy, F1 Score is also calculated for each trained model on the test set. Then, $F1_{robust}$ is used as the overall performance measure, similar to the definition of $ACC_{robust}$ in equation (8).

The evaluation results of the MLP and CNN models on the MIT-BIH dataset and the CPSC2018 dataset are visualized by Fig.3 and reported in Table 4 in Appendix. Adversarial examples are shown in Fig. 4 and 5. All figures are in high-resolution, please zoom in for better visualization.

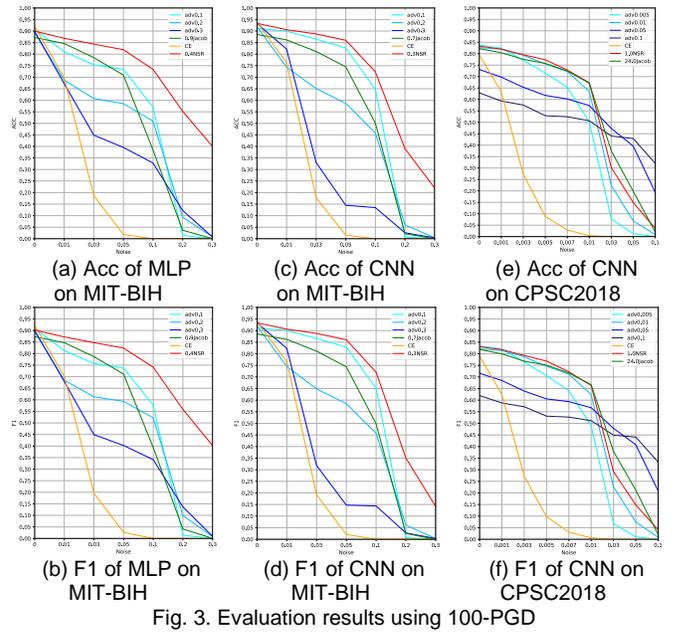

(a) Acc of MLP on MIT-BIH
(b) F1 of MLP on MIT-BIH
(c) Acc of CNN on MIT-BIH
(d) F1 of CNN on MIT-BIH
(e) Acc of CNN on CPSC2018
(f) F1 of CNN on CPSC2018

Fig. 3. Evaluation results using 100-PGD

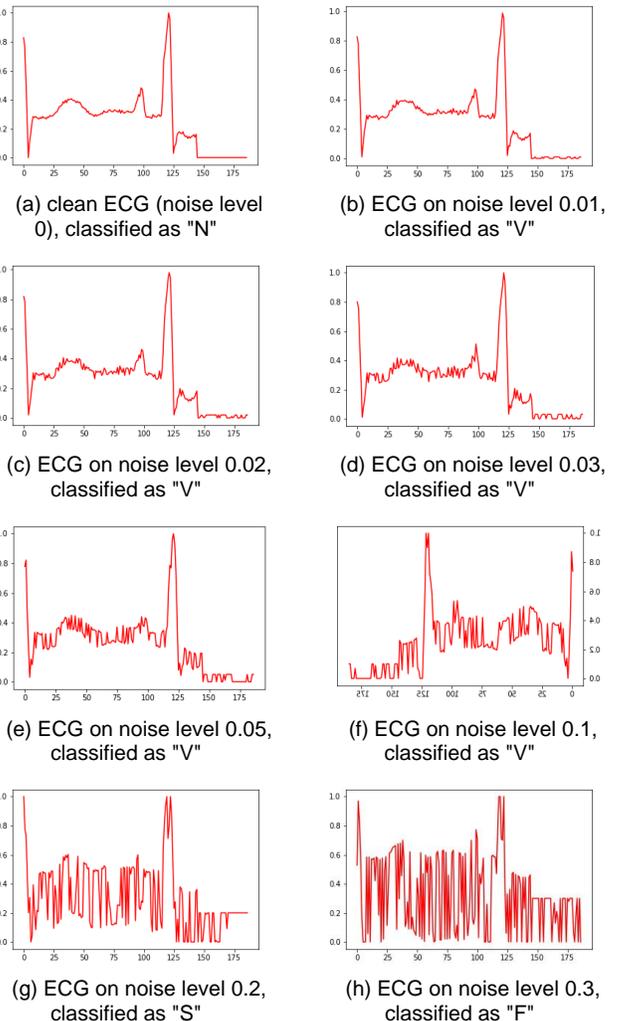

(a) clean ECG (noise level 0), classified as "N"
(b) ECG on noise level 0.01, classified as "V"
(c) ECG on noise level 0.02, classified as "V"
(d) ECG on noise level 0.03, classified as "V"
(e) ECG on noise level 0.05, classified as "V"
(f) ECG on noise level 0.1, classified as "V"
(g) ECG on noise level 0.2, classified as "S"
(h) ECG on noise level 0.3, classified as "F"

Fig. 4. MIT-BIT ECG signals on different noise levels. The original classification label is N (normal). However, when the noise level is 0.01, the MLP classified it as V (premature ventricular contraction and ventricular escape).

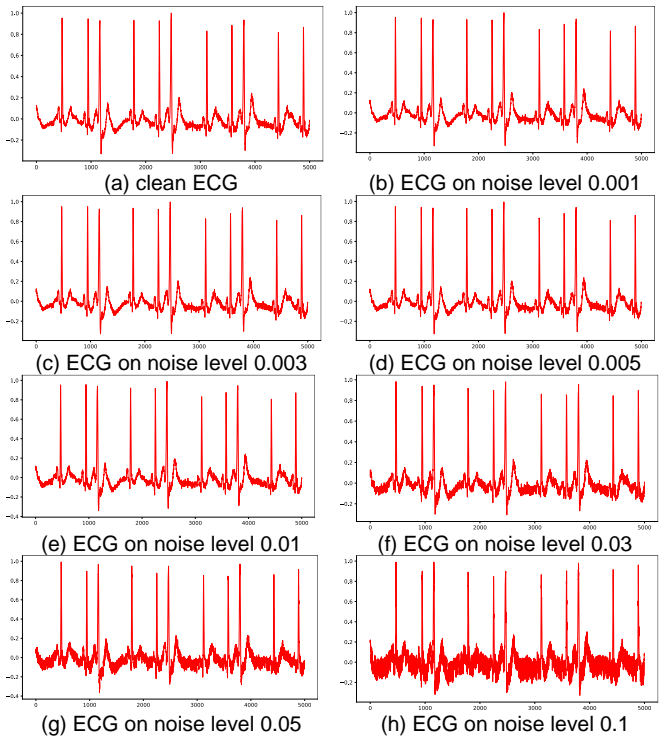

Fig. 5. CPSC2018 ECG Lead I on different noise levels

Our method achieved the best accuracy and F1 score on all noise levels under the 100-PGD attack to the MLP on MIT-BIH test set (Fig. 3(a)-(b)), On the noise level of 0.1, the MLP model "0.4NSR" has an accuracy of approximately 75%, which is the highest among the trained models.

Our method achieved the best accuracy and F1 score when the noise level is within 0.1 under the 100-PGD attack to the CNN on MIT-BIH test set (Fig. 3(c)-(d)). On the noise level of 0.1, the CNN model "0.3NSR" can maintain an accuracy of more than 71%. "adv0.2" and "adv0.3 are not as good as 0.3NSR on all of the noise levels, which means these adversarially-trained models are not strong enough to resist against the 100-PGD attack.

Our method performed the best within the noise level of 0.01 under the 100-PGD attack to the CNN on CPSC2018 test set (Fig. 3(e)-(f)), and the other two methods also improved the robustness of the CNN against 100-PGD adversarial attack. The 100-PGD attack is so strong that the accuracy of the model "CE" dropped close to 0% on the noise level of 0.01. As a comparison, on the noise level of 0.01, the three CNN models, "1.0NSR", "24.0Jacob" and "adv0.01" kept an average accuracy of more than 60% and F1 score of more than 0.6. Compared to the model "CE", the increase in clean accuracy of the model "1.0NSR" is achieved by using the NSR Regularization, and the increase in clean accuracy of the model ""adv0.01" is achieved by data augmentation with adversarial samples. This improvement is significant in this challenging task, considering the small sample size (5905 for training) and high input dimension (33792 timepoints per signal). The results also suggest that improving robustness may not always be in conflict of increasing accuracy, as compared to the results on the MIT-BIH dataset. However, we observed that the "adv $\epsilon$" modes have some weaknesses. First, it is very sensitive to the user-defined noise level $\epsilon$, which makes the model to be robust only around the specified noise level of $\epsilon$. From Fig. 3(e) and (f), the model "adv0.01" has a good resist to 100-PGD on the noise levels of no larger than 0.01, and its accuracy dropped significantly when the noise level becomes higher. For the model "adv0.05", the accuracy dropped slowly before the noise level of 0.05 and then dropped sharply. Also, a very large noise $\epsilon$ for adversarial training led to a significant reduction of classification accuracy on clean data (e.g., "adv0.05" and "adv0.1"). This can be explained from the preceptive of decision boundary: a very large perturbation on the clean sample can push the input sample to cross the optimal decision boundary for classification, and as a result, training with this adversarial sample will lead to a decrease in accuracy. A very small noise $\epsilon$ for adversarial training has almost no effect on CNN robustness (e.g., "adv0.005").

We also did an informal comparison with the ECG classification methods reported in the CPSC2018 competition [24]. The model "CE" that was trained only with the standard cross-entropy loss, achieved an accuracy of ~80% and F1 score of ~0.80 on clean data, which would rank in top 6 (F1 scores of the Top-6 methods are 0.837, 0.830, 0.806, 0.802, 0.791 and 0.783). The two models "1.0NSR" and "adv0.01" had higher accuracy of ~ 83% and F1 score of ~ 0.83 on clean data, which would rank in top-3 in the CPSC2018 competition. We note that the top-3 methods in the CPSC2018 competition used recurrent neural networks. Our results show that a similar accuracy for this classification task on clean data can be achieved by a CNN without recurrent layers.

*D. Evaluation of the Methods using the 100-SAP Attack*

We evaluated the methods using the SAP adversarial attack [14] with different noise levels. The SAP attack was designed specifically to attack ECG signal classifiers by generating smoothed adversarial ECG signals.

In this experiment, we used the source code provided by the authors of the SAP attack [14]. We name it 100-SAP because we set the number of iterations in SAP to 100, similar to the settings of 100-PGD attack. Other parameters of SAP take the default values. Compared to the 100-PGD attack, the 100-SAP attack is a weaker attack, as shown by the results in Fig. 6 and Table 5 in Appendix.

Our method achieved the best performance on all noise levels under the SAP attack to the MLP on the MIT-BIH dataset (Fig.6 (a)-(b)). On the noise level of 0.1, the MLP model "0.4NSR" can still hold an accuracy of over 80%., but the accuracy of the MLP model "0.9Jacob" dropped close to 50%, and all of the "adv $\epsilon$" models had relatively poor performance.

Our method achieved the best performance on all noise levels under the SAP attack to the CNN on MIT-BIH test set (Fig.6 (c)-(d)). On the noise level of 0.1, the MLP model "0.4NSR" could still hold an accuracy close to 85%., and the accuracy of the MLP model "0.7Jacob" dropped close to 65%. Among the "adv" models, only "adv0.1" had a comparable performance.

Our method achieved the best performance within the noise level of 0.01 under the SAP attack to the CNN on CPSC2018 test set (Fig.6 (e)-(f)). When the noise level reached 0.01, the models "1.0NSR" and "24.0Jacob" could still hold an accuracy over 80%. Among the "adv ϵ" models, only "adv0.01" had comparable performance. Compared to the results under the 100-PGD attack, the two models "1.0NSR" and "24.0Jacob" had much better performance, which indicates the SPA attack is much weaker than the 100-PGD attack.

good at resisting the adversarial noises generated from the same type of adversarial attack, and the regularization-based methods are not tied to a specific type of attack.

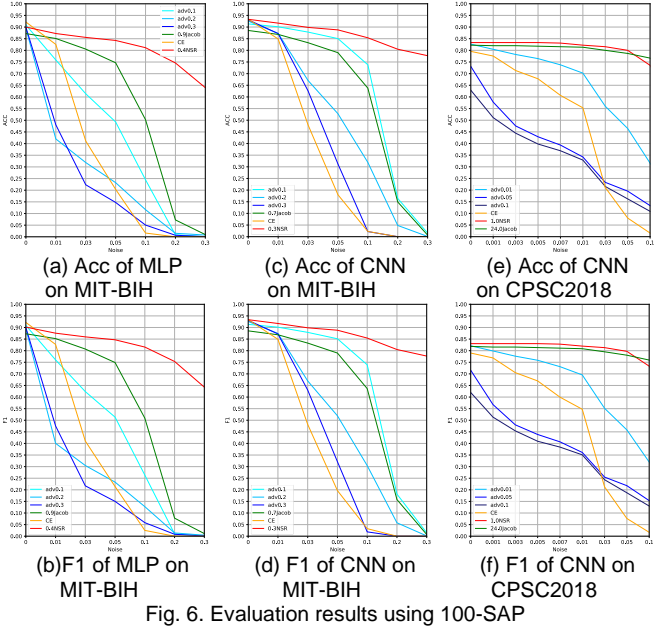

Fig. 6. Evaluation results using 100-SAP

(a) Acc of MLP on MIT-BIH
(b) F1 of MLP on MIT-BIH
(c) Acc of CNN on MIT-BIH
(d) F1 of CNN on MIT-BIH
(e) Acc of CNN on CPSC2018
(f) F1 of CNN on CPSC2018

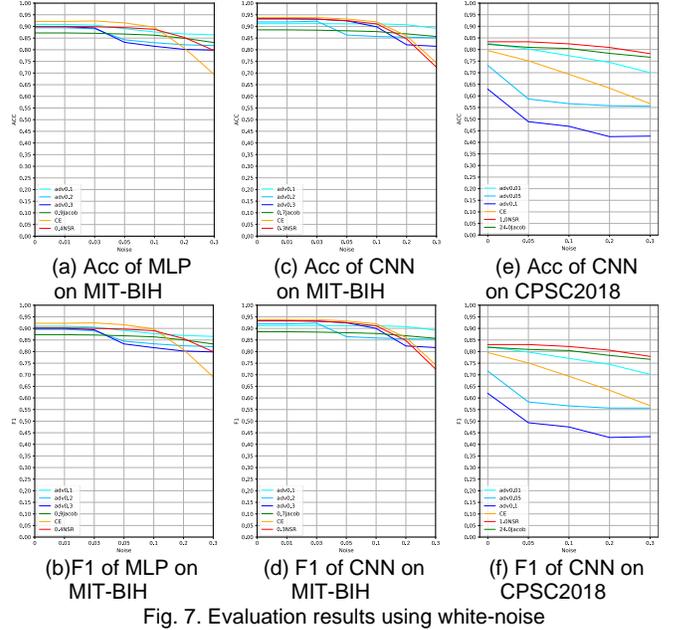

Fig. 7. Evaluation results using white-noise

(a) Acc of MLP on MIT-BIH
(b) F1 of MLP on MIT-BIH
(c) Acc of CNN on MIT-BIH
(d) F1 of CNN on MIT-BIH
(e) Acc of CNN on CPSC2018
(f) F1 of CNN on CPSC2018

### E. Evaluation of the Methods using White-noise Attack

In addition, we tested the performance of each trained model under white-noises attacks, and the results are visualized by Fig. 7 and reported in Table 6 in Appendix. Basically, white noises from a uniform distribution were added to the samples in the test sets, and the accuracy of each trained model on these noisy samples were measured. The maximum amplitude of the white noises is the noise level that was varied in a large range. Compared with the 100-PGD attack and the 100-SAP attack, random white noises are more like the natural noises in ECG signals, which are originated from ECG recording device (e.g., electromyogram noise and power line interference).

Under the white-noise attack to the model (MLP and CNN) on MIT-BIH test set (Fig.7 (a)-(d)), the model "CE" trained with the standard cross-entropy performed the best within the noise level of 0.1, which indicates that the standard classifiers naturally have some resist to white noises. The other methods performed slightly worse, which indicates that robustness to adversarial noises comes with a small cost on the accuracy on white-noise corrupted signals.

Under the white-noise attack to the CNN on CPSC2018 test set (Fig.7 (e)-(f)), our method (i.e., "1.0NSR") performed the best within the noise level of 0.3, and the CNN model "CE" performed much worse. The model "24.0Jacob" performed closely to the model "1.0NSR". A rather surprising result is that the accuracy of the "adv ϵ" models dropped significantly as the white noise level increased, and the performance of the "adv ϵ" models was even not as good as that in the 100-PGD evaluation. The reason could be that an adversarially-trained model is only

## IV. DISCUSSION

Our preliminary study was presented at conferences [32, 33], which, to our best knowledge, is the first work to improve DNN robustness against adversarial noises for ECG signal classification. In this paper, we present our complete study, including (1) extensive evaluation of the proposed method on two datasets, (2) comparison with other methods with fine-tuned parameters, and (3) evaluation against the SAP attack and the white-noise attack. Besides our work, neither adversarial training nor Jacobian Regularization have been applied for ECG signal classification.

The fairness of method comparison is ensured by rigorous evaluations. We used the 100-PGD attack, which is widely used for adversarial robustness research [28, 30, 31], and used the 100-SAP attack that was designed specifically to generate smoothed adversarial ECG samples. Compared to other white-box adversarial attacks, such as DeepFool [32] and C&W attack [33] as summarized in [28], the noise level can be controlled in the PGD and SAP algorithms so that the robustness of a model under different noise levels can be investigated. After the best parameters are found through a series of training-validation experiments, the performance of every trained model was evaluated on the test sets.

In general, our method performed the best under the 100-PGD attack and the 100-SAP attack. The evaluation results on the test sets of the two datasets are summarized in the Table A and Table B, which also shows 100-SAP is weaker than 100-PGD. Under the white-noise attack, our method also achieved competitive performance, as summarized in Table C.

TABLE A
$ACC_{robust}$ of the best models under the 100-PGD attack

|  | NSR | Jacob | adv | CE |
|---|---|---|---|---|
| MLP on MIT-BIH | **85.60%** | 76.55% | 80.74% | 42.34% |
| CNN on MIT-BIH | **88.64%** | 79.94% | 85.66% | 43.57% |
| CNN on CPSC2018 | **79.67%** | 78.56% | 78.74% | 41.37% |

TABLE B
$ACC_{robust}$ of the best models under the 100-SAP attack

|  | NSR | Jacob | adv | CE |
|---|---|---|---|---|
| MLP on MIT-BIH | **87.19%** | 79.20% | 68.46% | 54.94% |
| CNN on MIT-BIH | **91.04%** | 82.93% | 87.52% | 56.23% |
| CNN on CPSC2018 | **83.22%** | 81.99% | 79.32% | 72.95% |

TABLE C
$ACC_{robust}$ of the best models under the withe-noise attack

|  | NSR | Jacob | adv | CE |
|---|---|---|---|---|
| MLP on MIT-BIH | 89.74% | 86.99% | 90.11% | **91.75%** |
| CNN on MIT-BIH | 92.88% | 88.35% | 92.65% | **93.47%** |
| CNN on CPSC2018 | **79.92%** | 79.48% | 75.40% | 67.95% |

By definition [26], adversarial noise is small in magnitude but leads to a large change in the model output. Thus, for method evaluation, we only need to focus on a limited range of noise levels. On the MIT-BIH dataset, when the noise level of the PGD-based adversarial attack is higher than 0.1, the adversarial samples/signals are hardly recognizable by the human eye (see Fig 4). On the CPSC2018 dataset, when the noise level is higher than 0.01, the noises in the adversarial samples/signals become noticeable by the human eye (see Fig.5), and the performance of the methods dropped sharply after this noise level. In other words, it is much harder to train a robust model for the CPSC2018 dataset.

The difficulty of improving model robustness on the CPSC2018 dataset is caused by data annotation/labeling and the sample size. The MIT-BIH dataset has been labeled beat-by-beat, and therefore, we could do classification on the beat level with plenty of training samples (87554). The CPSC2018 dataset only has class labels on the recording level, and therefore there is a limited number of training samples (5905). In general, a DNN model trained with a larger number of samples will have a higher accuracy. Our recent work shows that training set size may affect adversarial robustness [34]: a larger training set may lead to a higher robustness. Thus, robustness could be further improved on the CPSC2018 dataset if the ECG recordings could be annotated beat-by-beat by human-expert cardiologists.

## V. CONCLUSION

In this paper, we proposed a novel method to improve the robustness of deep neural networks for classification of ECG signals. Our method aims to reduce the proportion of introduced noises in the output of the neural network, namely, noise-to-signal ratio (NSR). In this way, our proposed method can help reduce the effect of noise on the prediction of the neural network and thus improve the robustness against adversarial attacks. The results of the experiments have demonstrated that our proposed NSR regularization-based method, in general, outperforms the other two methods under the PGD and SAP adversarial attacks for the classification task. We hope that our approach may facilitate the development of robust solutions for automated ECG diagnosis.

## APPENDIX

### DATA PREPARATION ON CPSC2018 DATASET

First, we removed the samples with multiple labels (477 samples). Second, from each class, we randomly put 5 samples into the validation set and 50 samples into the test set. In this way, we split the dataset into a training set of 5905 samples, a validation set of 45 samples, and a test set of 450 samples. Because this dataset is class-imbalanced, upsampling was done to ensure that the training set is balanced in the number of samples in different categories. Third, for each sample, we removed Lead 3, 4, 5 and 6, which is due to the fact that Lead 3, 4, 5 and 6 are redundant and can be derived from other leads [30]. Fourth, for each sample, we scaled each lead by its maximum absolute value to make each lead to be within the range of -1 to 1 [31], which is the standard way to do feature normalization. Then, we randomly padded each sample with 0 on both ends during training, such that the temporal length of a padded single is 33792. For a few samples with a length larger than 33792, we discarded the signal elements after 33792. As a result, each sample after preprocessing is a tensor with the shape of 8 × 33792. Randomly padding 0 on both ends of a signal is data augmentation to shift a single to left or right along the time axis. We have made the preprocessing method publicly available (see the link at the end of the paper).

### TIME COST OF MODEL TRAINING

The experiments were conducted on a server with Nvidia Tesla V100 GPU (32 GB memory) and Intel(R) Xeon(R) E5-2698 v4 CPU (2.20GHz). The operating system is CentOS 7. Python version is 3.8.3, and Pytorch version is 1.5.0. On the CPSC2018 dataset, it took about 2 hours to train a "CE" model, 12 hours to train a "βNSR" model, 24 hours to train a "λJacob" model, and 12 hours to train an "adv ϵ" model. On the BIT-MIT dataset, for the CNN, it took about 1.28 hours to train a "CE" model, 8.89 hours to train a "βNSR" model, 8.61 hours to train a " λJacob " model, and 9.17 hours to train an "adv ϵ" model. On the BIT-MIT dataset, for the MLP, it took about 0.36 hour to train a "CE" model, 1.81 hours to train a "βNSR" model, 1.15 hours to train a " λJacob " model, and 1.31 hours to train an "adv ϵ" model.

### TABLE 1

A to C show the accuracy of each "βNSR" model (MLP or CNN) on validation set under the 100-PGD attack with noise levels in {0.01, 0.03, 0.05, 0.1, 0.2, 0.3}. Clean data has the noise level of 0.

TABLE 1 A.
Accuracy of MLP "βNSR" on MIT-BIT validation set

|  | 0 | 0.01 | 0.03 | 0.05 | 0.1 | 0.2 | 0.3 | $ACC_{robust}$ |
|---|---|---|---|---|---|---|---|---|
| 0.1NSR | 89.59% | 84.31% | 73.74% | 63.28% | 42.07% | 18.10% | 11.26% | 76.04% |
| 0.2NSR | 90.47% | 86.16% | 82.39% | 77.91% | 67.42% | 45.39% | 25.61% | 84.03% |
| 0.3NSR | 90.36% | 84.63% | 76.08% | 64.99% | 48.02% | 30.69% | 19.75% | 77.91% |
| 0.4NSR | 90.18% | 86.06% | 83.72% | 81.26% | 73.30% | 55.80% | 40.36% | 85.43% |
| 0.5NSR | 90.55% | 85.41% | 77.36% | 71.82% | 59.32% | 36.31% | 25.01% | 81.18% |
| 0.6NSR | 91.00% | 85.74% | 77.85% | 66.74% | 50.23% | 29.96% | 16.74% | 79.18% |
| 0.7NSR | 91.32% | 83.73% | 71.18% | 59.85% | 39.89% | 17.25% | 7.33% | 75.42% |
| 0.8NSR | 90.63% | 84.35% | 79.59% | 74.47% | 59.16% | 32.20% | 20.36% | 81.87% |
| 0.9NSR | 90.44% | 82.79% | 65.26% | 55.00% | 48.75% | 38.08% | 29.10% | 74.54% |

### TABLE 1 B.
### Accuracy of CNN "βNSR" on MIT-BIT validation set

|  | 0 | 0.01 | 0.03 | 0.05 | 0.1 | 0.2 | 0.3 | $ACC_{robust}$ |
|---|---|---|---|---|---|---|---|---|
| 0.1NSR | 94.10% | 91.17% | 88.96% | 85.09% | 68.10% | 32.80% | 20.45% | 88.36% |
| 0.2NSR | 93.67% | 88.58% | 82.58% | 73.90% | 58.37% | 30.74% | 15.14% | 83.79% |
| 0.3NSR | 93.79% | 91.62% | 89.53% | 87.08% | 73.65% | 39.44% | 22.03% | 89.40% |
| 0.4NSR | 94.87% | 91.47% | 86.08% | 77.30% | 53.03% | 32.78% | 19.52% | 84.91% |
| 0.5NSR | 93.62% | 89.87% | 86.27% | 80.92% | 60.20% | 27.96% | 17.73% | 85.88% |
| 0.6NSR | 93.95% | 91.72% | 88.69% | 84.87% | 69.24% | 40.61% | 25.07% | 88.41% |
| 0.7NSR | 93.87% | 90.72% | 89.29% | 86.91% | 71.77% | 41.24% | 23.56% | 89.07% |
| 0.8NSR | 92.95% | 88.10% | 82.48% | 73.56% | 57.07% | 28.18% | 17.13% | 83.14% |
| 0.9NSR | 92.53% | 89.00% | 85.99% | 82.94% | 68.56% | 39.39% | 23.63% | 86.76% |

### TABLE 1 C.
### Accuracy of CNN "βNSR" on CPSC2018 validation set

|  | 0 | 0.001 | 0.003 | 0.005 | 0.007 | 0.01 | 0.03 | 0.05 | 0.1 | $ACC_{robust}$ |
|---|---|---|---|---|---|---|---|---|---|---|
| 0.4NSR | 73.33% | 73.33% | 73.33% | 71.11% | 66.67% | 55.56% | 6.67% | 2.22% |  | 70.90% |
| 0.6NSR | 77.78% | 77.78% | 77.78% | 75.56% | 73.33% | 62.22% | 24.44% | 11.11% | 0.00% | 75.81% |
| 0.8NSR | 77.78% | 77.78% | 77.78% | 77.78% | 73.33% | 64.44% | 31.11% | 15.56% | 6.67% | 76.21% |
| 0.9NSR | 73.33% | 73.33% | 73.33% | 68.89% | 62.22% | 53.33% | 26.67% | 8.89% | 0.00% | 69.92% |
| 1.0NSR | 93.33% | 88.89% | 86.67% | 86.67% | 84.44% | 73.33% | 26.67% | 6.67% | 0.00% | 88.95% |
| 1.1NSR | 66.67% | 66.67% | 64.44% | 62.22% | 62.22% | 53.33% | 22.22% | 15.56% | 2.22% | 64.41% |
| 1.2NSR | 37.78% | 37.78% | 37.78% | 37.78% | 37.78% | 37.78% | 35.56% | 35.56% | 17.78% | 37.78% |

### TABLE 2

A to C show the accuracy of each "λJacob" model on validation set under the 100-PGD attack with noise levels in {0.01, 0.03, 0.05, 0.1, 0.2, 0.3}. Clean data has the noise level of 0.

### TABLE 2 A.
### Accuracy of MLP "λJacob" on MIT-BIT validation set

|  | 0 | 0.01 | 0.03 | 0.05 | 0.1 | 0.2 | 0.3 | $ACC_{robust}$ |
|---|---|---|---|---|---|---|---|---|
| 0.5Jacob | 90.09% | 87.44% | 78.26% | 68.63% | 26.88% | 1.68% | 0.02% | 75.94% |
| 0.6Jacob | 89.78% | 87.47% | 79.69% | 70.62% | 30.61% | 2.28% | 0.02% | 76.93% |
| 0.7Jacob | 89.11% | 86.35% | 79.40% | 70.40% | 32.38% | 3.46% | 0.03% | 76.70% |
| 0.8Jacob | 88.60% | 86.45% | 78.67% | 70.64% | 31.50% | 3.33% | 0.07% | 76.31% |
| 0.9Jacob | 88.27% | 85.63% | 79.35% | 71.23% | 36.62% | 3.58% | 0.05% | 77.03% |
| 1.0Jacob | 87.09% | 84.91% | 78.99% | 71.15% | 38.91% | 6.10% | 0.39% | 76.68% |
| 1.1Jacob | 86.30% | 84.51% | 78.77% | 71.72% | 43.09% | 4.40% | 0.13% | 76.95% |
| 1.2Jacob | 85.66% | 83.55% | 78.07% | 71.71% | 43.61% | 7.18% | 0.50% | 76.56% |
| 1.3Jacob | 85.54% | 83.48% | 78.10% | 72.02% | 44.26% | 7.56% | 0.55% | 76.66% |
| 1.4Jacob | 85.15% | 82.68% | 77.14% | 71.56% | 44.59% | 6.32% | 0.28% | 76.25% |
| 1.5Jacob | 84.94% | 82.96% | 77.24% | 71.92% | 46.70% | 7.96% | 0.78% | 76.55% |

### TABLE 2 B.
### Accuracy of CNN "λJacob" on MIT-BIT validation set

|  | 0 | 0.01 | 0.03 | 0.05 | 0.1 | 0.2 | 0.3 | $ACC_{robust}$ |
|---|---|---|---|---|---|---|---|---|
| 0.5Jacob | 89.99% | 87.99% | 82.44% | 74.53% | 41.60% | 1.41% | 0.01% | 79.75% |
| 0.6Jacob | 89.41% | 87.64% | 81.86% | 74.08% | 45.07% | 1.41% | 0.01% | 79.78% |
| 0.7Jacob | 89.08% | 86.96% | 81.85% | 74.37% | 51.79% | 2.05% | 0.01% | 80.56% |
| 0.8Jacob | 88.99% | 86.55% | 81.87% | 74.91% | 51.39% | 3.91% | 0.18% | 80.53% |
| 0.9Jacob | 87.72% | 86.14% | 81.50% | 75.22% | 52.21% | 5.93% | 0.02% | 80.02% |
| 1.0Jacob | 87.81% | 86.12% | 81.58% | 75.07% | 53.00% | 4.74% | 0.06% | 80.14% |
| 1.1Jacob | 87.97% | 85.74% | 81.01% | 75.07% | 50.33% | 4.52% | 0.06% | 79.76% |
| 1.2Jacob | 86.85% | 84.84% | 80.46% | 75.40% | 50.23% | 4.04% | 0.06% | 79.14% |
| 1.3Jacob | 86.23% | 84.78% | 80.49% | 75.85% | 56.51% | 6.68% | 0.37% | 79.77% |
| 1.4Jacob | 86.45% | 85.06% | 80.93% | 75.93% | 57.40% | 6.94% | 0.25% | 80.09% |
| 1.5Jacob | 86.07% | 84.56% | 80.77% | 75.93% | 54.79% | 6.16% | 0.04% | 79.49% |

### TABLE 2 C.
### Accuracy of CNN "λJacob" on CPSC2018 validation set

|  | 0 | 0.001 | 0.003 | 0.005 | 0.007 | 0.01 | 0.03 | 0.05 | 0.1 | $ACC_{robust}$ |
|---|---|---|---|---|---|---|---|---|---|---|
| 4.0Jacob | 86.67% | 82.22% | 77.78% | 73.33% | 62.22% | 53.33% | 11.11% | 2.22% | 0.00% | 78.14% |
| 14.0Jacob | 86.67% | 82.22% | 82.22% | 75.56% | 68.89% | 62.22% | 24.44% | 20.00% | 0.00% | 80.50% |
| 24.0Jacob | 86.67% | 86.67% | 80.00% | 80.00% | 80.00% | 75.56% | 31.11% | 17.78% | 2.22% | 83.61% |
| 34.0Jacob | 84.44% | 82.22% | 80.00% | 80.00% | 75.56% | 71.11% | 35.56% | 20.00% | 0.00% | 81.22% |
| 44.0Jacob | 86.67% | 84.44% | 80.00% | 77.78% | 71.11% | 68.89% | 44.44% | 28.89% | 2.22% | 81.51% |
| 54.0Jacob | 82.22% | 80.00% | 80.00% | 77.78% | 75.56% | 75.56% | 44.44% | 24.44% | 0.00% | 80.03% |
| 64.0Jacob | 82.22% | 82.22% | 82.22% | 77.78% | 71.11% | 68.89% | 40.00% | 26.67% | 4.44% | 79.34% |
| 74.0Jacob | 80.00% | 80.00% | 77.78% | 75.56% | 73.33% | 71.11% | 46.67% | 28.89% | 4.44% | 77.80% |
| 84.0Jacob | 73.33% | 73.33% | 73.33% | 71.11% | 68.89% | 66.67% | 35.56% | 31.11% | 2.22% | 72.04% |

### TABLE 3

A to C show the accuracy of each "adv ε" model on validation set under the 100-PGD attack noise levels in {0.01, 0.03, 0.05, 0.1, 0.2, 0.3}. Clean data has the noise level of 0.

### TABLE 3 A.
### Accuracy of MLP "adv ε" on MIT-BIT validation set

|  | 0 | 0.01 | 0.03 | 0.05 | 0.1 | 0.2 | 0.3 | $ACC_{robust}$ |
|---|---|---|---|---|---|---|---|---|
| adv0.05 | 92.60% | 90.73% | 85.09% | 75.22% | 23.47% | 0.05% | 0.00% | 79.03% |
| adv0.15 | 90.04% | 71.24% | 65.75% | 64.77% | 59.14% | 3.94% | 0.07% | 76.97% |
| adv0.1 | 91.82% | 80.13% | 71.85% | 69.57% | 57.18% | 1.62% | 0.03% | 79.95% |
| adv0.25 | 89.37% | 60.07% | 44.60% | 45.63% | 43.65% | 9.76% | 1.00% | 66.36% |
| adv0.2 | 89.75% | 62.46% | 48.41% | 47.95% | 47.70% | 8.32% | 0.73% | 68.47% |
| adv0.3 | 90.50% | 58.36% | 41.74% | 40.07% | 38.78% | 15.76% | 1.60% | 64.06% |

### TABLE 3 B.
### Accuracy of CNN "adv ε" on MIT-BIT validation set

|  | 0 | 0.01 | 0.03 | 0.05 | 0.1 | 0.2 | 0.3 | $ACC_{robust}$ |
|---|---|---|---|---|---|---|---|---|
| adv0.05 | 93.36% | 91.88% | 87.95% | 82.31% | 30.19% | 0.07% | 0.00% | 82.21% |
| adv0.15 | 93.41% | 81.89% | 79.00% | 75.19% | 62.34% | 3.21% | 0.20% | 83.51% |
| adv0.1 | 91.80% | 90.50% | 87.44% | 83.28% | 64.93% | 1.05% | 0.11% | 86.25% |
| adv0.25 | 94.85% | 80.71% | 35.16% | 18.10% | 6.65% | 4.12% | 0.98% | 54.99% |
| adv0.2 | 93.68% | 83.91% | 63.99% | 57.47% | 40.75% | 7.41% | 0.30% | 75.20% |
| adv0.3 | 94.22% | 77.79% | 25.90% | 7.15% | 6.92% | 1.07% | 0.49% | 49.29% |

### TABLE 3 C.
### Accuracy of CNN "adv ε" on CPSC2018 validation set

|  | 0 | 0.001 | 0.003 | 0.005 | 0.007 | 0.01 | 0.03 | 0.05 | 0.1 | $ACC_{robust}$ |
|---|---|---|---|---|---|---|---|---|---|---|
| adv0.005 | 82.22% | 80.00% | 75.56% | 68.89% | 62.22% | 48.89% | 8.89% | 0.00% | 0.00% | 74.71% |
| adv0.01 | 84.44% | 84.44% | 80.00% | 73.33% | 71.11% | 64.44% | 13.33% | 2.22% | 0.00% | 79.58% |
| adv0.02 | 84.44% | 80.00% | 75.56% | 73.33% | 66.67% | 62.22% | 31.11% | 6.67% | 0.00% | 77.97% |
| adv0.03 | 75.56% | 75.56% | 73.33% | 68.89% | 68.89% | 68.89% | 42.22% | 17.78% | 2.22% | 73.30% |
| adv0.04 | 68.89% | 64.44% | 62.22% | 55.56% | 57.78% | 55.56% | 42.22% | 31.11% | 4.44% | 63.99% |
| adv0.05 | 62.22% | 60.00% | 55.56% | 53.33% | 57.78% | 53.33% | 40.00% | 28.89% | 13.33% | 59.20% |
| adv0.06 | 68.89% | 62.22% | 62.22% | 53.33% | 55.56% | 51.11% | 42.22% | 35.56% | 17.78% | 62.91% |
| adv0.07 | 57.78% | 53.33% | 48.89% | 44.44% | 44.44% | 44.44% | 37.78% | 28.89% | 17.78% | 52.30% |
| adv0.08 | 60.00% | 60.00% | 51.11% | 51.11% | 46.67% | 48.89% | 37.78% | 35.56% | 24.44% | 55.56% |
| adv0.09 | 62.22% | 55.56% | 53.33% | 46.67% | 42.22% | 48.89% | 37.78% | 35.56% | 24.44% | 55.40% |
| adv0.1 | 42.22% | 42.22% | 37.78% | 35.56% | 35.56% | 33.33% | 26.67% | 26.67% | 20.00% | 39.52% |

### TABLE 4

A to F show the Accuracy and F1 of the models on test set under the 100-PGD attack with noise levels noise levels in {0.01, 0.03, 0.05, 0.1, 0.2, 0.3}. Clean data has the noise level of 0.

### TABLE 4 A.
### Accuracy of MLP models on MIT-BIT test set

|  | 0 | 0.01 | 0.03 | 0.05 | 0.1 | 0.2 | 0.3 | $ACC_{robust}$ |
|---|---|---|---|---|---|---|---|---|
| adv0.1 | 90.87% | 81.04% | 75.43% | 73.40% | 57.07% | 1.41% | 0.01% | 80.74% |
| adv0.2 | 89.42% | 68.73% | 60.79% | 58.55% | 50.95% | 9.39% | 0.87% | 73.35% |
| adv0.3 | 89.91% | 67.21% | 44.89% | 39.56% | 32.82% | 12.29% | 0.84% | 64.03% |
| 0.9Jacob | 87.20% | 84.60% | 78.50% | 70.94% | 38.48% | 3.69% | 0.07% | 76.55% |
| CE | 92.16% | 69.74% | 18.65% | 1.86% | 0.00% | 0.00% | 0.00% | 42.34% |
| 0.4NSR | 89.99% | 86.86% | 84.35% | 81.92% | 73.43% | 55.27% | 40.14% | 85.60% |

### TABLE 4 B.
### F1 of MLP models on MIT-BIT test set

|  | 0 | 0.01 | 0.03 | 0.05 | 0.1 | 0.2 | 0.3 | $F1_{robust}$ |
|---|---|---|---|---|---|---|---|---|
| adv0.1 | 90.93% | 81.28% | 75.77% | 73.76% | 57.67% | 1.42% | 0.01% | 80.98% |
| adv0.2 | 89.55% | 68.46% | 61.32% | 59.34% | 52.21% | 9.94% | 0.80% | 73.81% |
| adv0.3 | 90.04% | 67.95% | 44.93% | 40.25% | 34.12% | 13.69% | 0.92% | 64.57% |
| 0.9Jacob | 87.29% | 84.72% | 78.66% | 71.15% | 39.34% | 4.10% | 0.06% | 76.79% |
| CE | 92.19% | 70.14% | 19.58% | 2.63% | 0.01% | 0.00% | 0.00% | 42.91% |
| 0.4NSR | 90.16% | 87.21% | 84.75% | 82.38% | 74.11% | 55.88% | 40.23% | 85.93% |

### TABLE 4 C.
### Accuracy of CNN models on MIT-BIT test set

|  | 0 | 0.01 | 0.03 | 0.05 | 0.1 | 0.2 | 0.3 | $ACC_{robust}$ |
|---|---|---|---|---|---|---|---|---|
| adv0.1 | 91.17% | 89.96% | 86.47% | 82.69% | 64.74% | 0.65% | 0.06% | 85.66% |
| adv0.2 | 91.91% | 74.36% | 65.09% | 58.69% | 45.75% | 6.03% | 0.40% | 74.72% |
| adv0.3 | 93.23% | 82.22% | 33.03% | 14.49% | 13.48% | 2.58% | 0.40% | 54.66% |
| 0.7Jacob | 88.53% | 86.17% | 81.15% | 74.52% | 50.11% | 2.20% | 0.00% | 79.94% |
| CE | 93.83% | 76.34% | 17.71% | 1.56% | 0.00% | 0.00% | 0.00% | 43.57% |
| 0.3NSR | 93.33% | 90.58% | 88.75% | 86.00% | 72.36% | 38.99% | 22.02% | 88.64% |

TABLE 4 D.
F1 of CNN models on MIT-BIT test set

|  | 0 | 0.01 | 0.03 | 0.05 | 0.1 | 0.2 | 0.3 | $F1_{robust}$ |
|---|---|---|---|---|---|---|---|---|
| adv0.1 | 91.21% | 90.02% | 86.59% | 82.85% | 65.13% | 0.65% | 0.05% | 85.77% |
| adv0.2 | 91.93% | 74.52% | 64.83% | 58.48% | 45.73% | 6.11% | 0.37% | 74.67% |
| adv0.3 | 93.24% | 82.34% | 31.83% | 14.73% | 14.37% | 2.73% | 0.34% | 54.73% |
| 0.7Jacob | 88.55% | 86.20% | 81.13% | 74.38% | 49.81% | 2.52% | 0.00% | 79.88% |
| CE | 93.83% | 76.54% | 19.19% | 2.08% | 0.00% | 0.00% | 0.00% | 44.11% |
| 0.3NSR | 93.36% | 90.61% | 88.77% | 86.03% | 72.07% | 35.05% | 14.20% | 88.63% |

TABLE 4 E.
Accuracy of CNN models on CPSC2018 test set

|  | 0 | 0.001 | 0.003 | 0.005 | 0.007 | 0.01 | 0.03 | 0.05 | 0.1 | $ACC_{robust}$ |
|---|---|---|---|---|---|---|---|---|---|---|
| adv0.005 | 83.78% | 82.44% | 77.11% | 71.56% | 65.33% | 50.22% | 7.78% | 1.33% | 0.00% | 76.66% |
| adv0.01 | 82.67% | 80.44% | 79.33% | 75.56% | 72.22% | 63.56% | 22.44% | 6.89% | 0.89% | 78.74% |
| adv0.05 | 73.11% | 69.78% | 65.33% | 61.78% | 60.22% | 57.33% | 47.33% | 39.56% | 19.33% | 67.98% |
| adv0.1 | 62.89% | 59.33% | 57.56% | 52.89% | 52.44% | 50.67% | 44.00% | 42.89% | 32.00% | 58.73% |
| CE | 79.56% | 63.33% | 27.33% | 8.89% | 2.89% | 0.44% | 0.00% | 0.00% | 0.00% | 41.37% |
| 1.0NSR | 83.33% | 82.00% | 79.56% | 77.33% | 72.89% | 67.33% | 30.22% | 14.89% | 3.56% | 79.67% |
| 24.0Jacob | 82.22% | 80.44% | 77.56% | 75.78% | 72.44% | 67.33% | 37.56% | 19.56% | 2.00% | 78.56% |

TABLE 4 F.
F1 of CNN models on CPSC2018 test set

|  | 0 | 0.001 | 0.003 | 0.005 | 0.007 | 0.01 | 0.03 | 0.05 | 0.1 | $F1_{robust}$ |
|---|---|---|---|---|---|---|---|---|---|---|
| adv0.005 | 83.33% | 82.05% | 76.49% | 70.58% | 64.08% | 48.93% | 6.76% | 1.00% | 0.00% | 75.96% |
| adv0.01 | 82.10% | 81.38% | 78.70% | 74.58% | 71.20% | 62.36% | 22.61% | 6.92% | 0.89% | 78.01% |
| adv0.05 | 71.58% | 68.45% | 63.92% | 60.54% | 59.37% | 56.75% | 47.86% | 40.78% | 20.80% | 66.67% |
| adv0.1 | 61.97% | 58.79% | 57.15% | 53.11% | 52.70% | 51.16% | 44.93% | 44.08% | 33.31% | 58.29% |
| CE | 78.99% | 62.40% | 27.23% | 9.74% | 3.12% | 0.60% | 0.00% | 0.00% | 0.00% | 41.28% |
| 1.0NSR | 83.04% | 81.78% | 79.27% | 76.83% | 72.23% | 66.32% | 29.15% | 14.80% | 4.12% | 79.26% |
| 24.0Jacob | 81.85% | 79.96% | 76.81% | 75.11% | 71.72% | 66.87% | 37.90% | 20.95% | 2.02% | 78.05% |

TABLE 5
A to F show the accuracy and F1 of the models on test set under the 100-SAP attack with noise levels in {0.01, 0.03, 0.05, 0.1, 0.2, 0.3}. Clean data has the noise level of 0.

TABLE 5 A.
Accuracy of MLP models on MIT-BIT test set

|  | 0 | 0.01 | 0.03 | 0.05 | 0.1 | 0.2 | 0.3 | $ACC_{robust}$ |
|---|---|---|---|---|---|---|---|---|
| adv0.1 | 90.87% | 75.88% | 61.39% | 49.30% | 24.47% | 0.88% | 0.07% | 68.46% |
| adv0.2 | 89.42% | 42.00% | 31.88% | 23.25% | 11.65% | 1.51% | 0.78% | 50.21% |
| adv0.3 | 89.91% | 48.08% | 22.35% | 14.75% | 5.04% | 0.67% | 0.10% | 45.08% |
| 0.9Jacob | 87.20% | 85.10% | 80.55% | 74.69% | 50.25% | 7.26% | 0.95% | 79.20% |
| CE | 92.16% | 82.56% | 41.13% | 20.37% | 1.62% | 0.00% | 0.00% | 54.94% |
| 0.4NSR | 89.99% | 87.23% | 85.57% | 84.26% | 81.15% | 74.60% | 64.01% | 87.19% |

TABLE 5 B.
F1 of MLP models on MIT-BIT test set

|  | 0 | 0.01 | 0.03 | 0.05 | 0.1 | 0.2 | 0.3 | $F1_{robust}$ |
|---|---|---|---|---|---|---|---|---|
| adv0.1 | 90.93% | 76.03% | 62.36% | 51.34% | 26.35% | 1.00% | 0.07% | 69.41% |
| adv0.2 | 89.55% | 40.05% | 30.48% | 23.32% | 12.75% | 1.40% | 0.44% | 50.01% |
| adv0.3 | 90.04% | 47.57% | 21.72% | 15.01% | 5.84% | 0.84% | 0.07% | 45.21% |
| 0.9Jacob | 87.29% | 85.21% | 80.71% | 74.89% | 50.90% | 7.78% | 1.14% | 79.40% |
| CE | 92.19% | 82.70% | 41.03% | 21.42% | 2.50% | 0.00% | 0.00% | 55.44% |
| 0.4NSR | 90.16% | 87.55% | 85.95% | 84.69% | 81.57% | 75.35% | 64.28% | 87.47% |

TABLE 5 C.
Accuracy of CNN models on MIT-BIT test set

|  | 0 | 0.01 | 0.03 | 0.05 | 0.1 | 0.2 | 0.3 | $ACC_{robust}$ |
|---|---|---|---|---|---|---|---|---|
| adv0.1 | 91.34% | 90.11% | 87.83% | 85.04% | 73.80% | 16.43% | 1.56% | 87.52% |
| adv0.2 | 92.56% | 87.27% | 67.05% | 52.79% | 31.90% | 4.83% | 0.12% | 73.01% |
| adv0.3 | 93.08% | 87.19% | 62.59% | 31.21% | 2.20% | 0.01% | 0.00% | 62.32% |
| 0.7Jacob | 88.53% | 86.83% | 83.27% | 78.99% | 63.77% | 14.92% | 0.68% | 82.93% |
| CE | 93.38% | 84.88% | 47.99% | 18.00% | 2.25% | 0.00% | 0.00% | 56.23% |
| 0.3NSR | 93.33% | 91.68% | 89.84% | 88.78% | 85.39% | 80.48% | 77.71% | 91.04% |

TABLE 5 D.
F1 of CNN models on MIT-BIT test set

|  | 0 | 0.01 | 0.03 | 0.05 | 0.1 | 0.2 | 0.3 | $F1_{robust}$ |
|---|---|---|---|---|---|---|---|---|
| adv0.1 | 91.36% | 90.14% | 87.90% | 85.14% | 73.91% | 18.16% | 1.47% | 87.57% |
| adv0.2 | 92.60% | 87.42% | 66.96% | 51.92% | 30.45% | 5.78% | 0.13% | 72.60% |
| adv0.3 | 93.09% | 87.22% | 63.02% | 32.11% | 1.87% | 0.01% | 0.00% | 62.57% |
| 0.7Jacob | 88.55% | 86.86% | 83.28% | 78.98% | 63.64% | 15.79% | 0.74% | 82.93% |
| CE | 93.38% | 84.91% | 48.22% | 19.66% | 3.26% | 0.00% | 0.00% | 56.96% |
| 0.3NSR | 93.36% | 91.73% | 89.86% | 88.83% | 85.45% | 80.50% | 77.68% | 91.08% |

TABLE 5 E.
Accuracy of CNN models on CPSC2018 test set

|  | 0 | 0.001 | 0.003 | 0.005 | 0.007 | 0.01 | 0.03 | 0.05 | 0.1 | $ACC_{robust}$ |
|---|---|---|---|---|---|---|---|---|---|---|
| adv0.01 | 82.67% | 80.44% | 78.22% | 76.44% | 73.78% | 70.22% | 56.00% | 46.44% | 31.56% | 79.32% |
| adv0.05 | 73.11% | 57.78% | 47.56% | 42.89% | 39.33% | 34.22% | 23.33% | 19.56% | 13.33% | 57.60% |
| adv0.1 | 62.89% | 51.11% | 44.44% | 39.78% | 36.89% | 32.89% | 21.56% | 16.22% | 10.89% | 51.28% |
| CE | 79.56% | 77.56% | 71.33% | 67.78% | 60.67% | 55.33% | 21.11% | 8.00% | 1.56% | 72.95% |
| 1.0NSR | 83.33% | 83.33% | 83.33% | 83.33% | 83.11% | 82.22% | 81.56% | 80.00% | 73.56% | 83.22% |
| 24.0Jacob | 82.22% | 82.00% | 82.00% | 81.78% | 81.56% | 81.33% | 80.00% | 78.67% | 76.67% | 81.99% |

TABLE 5 F.
F1 of CNN models on CPSC2018 test set

|  | 0 | 0.001 | 0.003 | 0.005 | 0.007 | 0.01 | 0.03 | 0.05 | 0.1 | $F1_{robust}$ |
|---|---|---|---|---|---|---|---|---|---|---|
| adv0.01 | 82.10% | 79.88% | 77.63% | 75.87% | 73.11% | 69.55% | 55.15% | 45.82% | 31.96% | 78.73% |
| adv0.05 | 71.58% | 56.76% | 48.02% | 43.87% | 40.70% | 36.12% | 25.49% | 21.78% | 15.34% | 57.42% |
| adv0.1 | 61.97% | 51.34% | 45.42% | 40.95% | 38.43% | 35.08% | 24.35% | 18.72% | 13.00% | 51.59% |
| CE | 78.99% | 76.87% | 70.53% | 66.96% | 59.97% | 54.75% | 21.26% | 7.63% | 1.68% | 72.30% |
| 1.0NSR | 83.04% | 83.04% | 83.04% | 83.04% | 82.82% | 81.98% | 81.31% | 79.71% | 73.31% | 82.93% |
| 24.0Jacob | 81.85% | 81.56% | 81.56% | 81.34% | 81.09% | 80.85% | 79.51% | 78.05% | 75.95% | 81.58% |

TABLE 6
A to F show the accuracy and F1 of the models on test set under the white noise attack with noise levels from 0 to 0.3.

TABLE 6 A.
Accuracy of MLP models on MIT-BIT test set

|  | 0 | 0.01 | 0.03 | 0.05 | 0.1 | 0.2 | 0.3 | $ACC_{robust}$ |
|---|---|---|---|---|---|---|---|---|
| adv0.1 | 90.87% | 90.87% | 90.54% | 89.04% | 87.62% | 86.79% | 86.38% | 90.11% |
| adv0.2 | 89.42% | 89.42% | 88.99% | 84.26% | 83.00% | 82.14% | 81.78% | 87.65% |
| adv0.3 | 89.91% | 89.91% | 89.34% | 83.18% | 81.49% | 80.15% | 79.83% | 87.59% |
| 0.9Jacob | 87.20% | 87.20% | 87.06% | 86.73% | 86.25% | 84.92% | 83.09% | 86.99% |
| CE | 92.16% | 92.16% | 92.34% | 91.52% | 89.67% | 80.73% | 69.36% | 91.75% |
| 0.4NSR | 89.99% | 89.99% | 89.84% | 89.49% | 88.77% | 85.27% | 79.62% | 89.74% |

TABLE 6 B.
F1 of MLP models on MIT-BIT test set

|  | 0 | 0.01 | 0.03 | 0.05 | 0.1 | 0.2 | 0.3 | $F1_{robust}$ |
|---|---|---|---|---|---|---|---|---|
| adv0.1 | 90.93% | 90.93% | 90.56% | 89.07% | 87.73% | 86.95% | 86.54% | 90.16% |
| adv0.2 | 89.55% | 89.55% | 89.00% | 84.50% | 83.36% | 82.50% | 82.14% | 87.82% |
| adv0.3 | 90.04% | 90.04% | 89.41% | 83.31% | 81.69% | 80.24% | 79.85% | 87.73% |
| 0.9Jacob | 87.29% | 87.29% | 87.15% | 86.82% | 86.36% | 85.05% | 83.26% | 87.08% |
| CE | 92.19% | 92.19% | 92.38% | 91.58% | 89.78% | 80.88% | 69.07% | 91.80% |
| 0.4NSR | 90.16% | 90.16% | 90.02% | 89.68% | 89.00% | 85.63% | 79.93% | 89.92% |

TABLE 6 C.
Accuracy of CNN models on MIT-BIT test set

|  | 0 | 0.01 | 0.03 | 0.05 | 0.1 | 0.2 | 0.3 | $ACC_{robust}$ |
|---|---|---|---|---|---|---|---|---|
| adv0.1 | 91.17% | 91.17% | 91.21% | 91.10% | 91.10% | 90.74% | 89.17% | 91.16% |
| adv0.2 | 91.91% | 91.91% | 92.23% | 86.30% | 85.69% | 85.37% | 85.11% | 90.17% |
| adv0.3 | 93.23% | 93.23% | 93.12% | 92.40% | 89.83% | 82.12% | 81.47% | 92.65% |
| 0.7Jacob | 88.55% | 88.55% | 88.39% | 88.12% | 87.79% | 86.81% | 85.68% | 88.35% |
| CE | 93.83% | 93.83% | 93.79% | 93.18% | 91.96% | 85.94% | 74.31% | 93.47% |
| 0.3NSR | 93.33% | 93.33% | 93.14% | 92.47% | 91.08% | 84.64% | 72.47% | 92.88% |

TABLE 6 D.
F1 of CNN models on MIT-BIT test set

|  | 0 | 0.01 | 0.03 | 0.05 | 0.1 | 0.2 | 0.3 | $F1_{robust}$ |
|---|---|---|---|---|---|---|---|---|
| adv0.1 | 91.21% | 91.21% | 91.24% | 91.13% | 91.13% | 90.78% | 89.24% | 91.19% |
| adv0.2 | 91.93% | 91.93% | 92.27% | 86.45% | 85.86% | 85.54% | 85.29% | 90.23% |
| adv0.3 | 93.24% | 93.24% | 93.13% | 92.43% | 90.00% | 82.43% | 81.74% | 92.68% |
| 0.7Jacob | 88.55% | 88.55% | 88.41% | 88.15% | 87.81% | 86.83% | 85.71% | 88.37% |
| CE | 93.83% | 93.83% | 93.79% | 93.18% | 91.99% | 86.13% | 74.35% | 93.48% |
| 0.3NSR | 93.36% | 93.36% | 93.17% | 92.51% | 91.12% | 84.79% | 72.46% | 92.91% |

TABLE 6 E.
Accuracy of CNN models on CPSC2018 test set

|  | 0 | 0.05 | 0.1 | 0.2 | 0.3 | $ACC_{robust}$ |
|---|---|---|---|---|---|---|
| adv0.01 | 82.67% | 80.22% | 77.33% | 74.44% | 70.00% | 75.40% |
| adv0.05 | 73.11% | 58.67% | 56.67% | 55.78% | 55.56% | 63.60% |
| adv0.1 | 62.89% | 48.89% | 46.89% | 42.44% | 42.67% | 51.95% |
| CE | 79.56% | 75.11% | 69.33% | 63.33% | 56.67% | 67.95% |
| 1.0NSR | 83.33% | 83.33% | 82.44% | 80.89% | 78.22% | 79.92% |
| 24.0Jacob | 82.22% | 80.93% | 80.45% | 78.34% | 76.68% | 79.48% |

TABLE 6 F.
F1 of CNN models on CPSC2018 test set

|  | 0 | 0.05 | 0.1 | 0.2 | 0.3 | $F1_{robust}$ |
|---|---|---|---|---|---|---|
| adv0.01 | 82.10% | 79.78% | 77.06% | 74.49% | 70.17% | 75.20% |
| adv0.05 | 71.58% | 58.28% | 56.58% | 55.64% | 55.63% | 62.91% |
| adv0.1 | 61.97% | 49.32% | 47.49% | 42.97% | 43.31% | 51.75% |
| CE | 79.56% | 75.11% | 69.33% | 63.33% | 56.67% | 67.95% |
| 1.0NSR | 83.04% | 83.04% | 82.19% | 80.61% | 77.91% | 79.50% |
| 24.0Jacob | 81.85% | 80.93% | 80.45% | 78.34% | 76.68% | 79.29% |

## DATA AVAILABILITY

The code of this study is publicly available at https://github.com/SarielMa/Robust_DNN_for_ECG

## COMPETING INTERESTS

The authors declare no competing interests.

## *AUTHOR CONTRIBUTIONS*

L. M. performed method development and experimental evaluation. L. L. designed the study. Both authors contribute to the writing of the manuscript and have read and approved the final manuscript.